\def\year{2021}\relax
\documentclass[letterpaper]{article} 
\usepackage{aaai21}  
\usepackage{times}  
\usepackage{helvet} 
\usepackage{courier}  
\usepackage[hyphens]{url}  
\usepackage{graphicx} 
\usepackage{amsmath}
\usepackage{multirow}
\usepackage{graphicx}
\usepackage{comment}
\usepackage{makecell}
\def\UrlFont{\rm}  
\usepackage{natbib}

\usepackage{caption} 
\usepackage{tabularx}
\usepackage{comment}
\usepackage{subcaption}
\usepackage[english]{babel}
\usepackage[autostyle]{csquotes}

\newcommand{\oneS}{\ensuremath{{}^{\textstyle *}}}
\newcommand{\twoS}{\ensuremath{{}^{\textstyle **}}}

\title{Arachnophobia Exposure Therapy using Experience-driven Procedural Content Generation via Reinforcement Learning (EDPCGRL)}
\author {
    Athar Mahmoudi-Nejad,
    Matthew Guzdial, 
    Pierre Boulanger  \\
}
\affiliations {
     Department of Computing Science, University of Alberta \\
    athar1@ualberta.ca, guzdial@ualberta.ca, pierreb@ualberta.ca
}

\begin{document}

\maketitle

\begin{abstract}
Personalized therapy, in which a therapeutic practice is adapted to an individual patient, leads to better health outcomes. 
Typically, this is accomplished by relying on a therapist's training and intuition along with feedback from a patient. 
While there exist approaches to automatically adapt therapeutic content to a patient, they rely on hand-authored, pre-defined rules, which may not generalize to all individuals.
In this paper, we propose an approach to automatically adapt therapeutic content to patients based on physiological measures. 
We implement our approach in the context of arachnophobia exposure therapy, and rely on experience-driven procedural content generation via reinforcement learning (EDPCGRL) to generate virtual spiders to match an individual patient. 
In this initial implementation, and due to the ongoing pandemic, we make use of virtual or artificial humans implemented based on prior arachnophobia psychology research.
Our EDPCGRL method is able to more quickly adapt to these virtual humans with high accuracy in comparison to existing, search-based EDPCG approaches. 
\end{abstract}

\section{Introduction}
Experience-driven Procedural Content Generation (EDPCG) is a PCG framework that modifies content to optimize a user's experience. 
Although EDPCG was developed for games, it can be applied to other HCI domains that require automated customized content (e.g. recommender systems) \cite{yannakakis2011experience}. 
We argue EDPCG can also be a useful framework for computer assisted-therapy.

We can divide computer-assisted therapy into two groups: non-adaptive and adaptive. 
The non-adaptive approach provides predesigned content for all users, which is not ideal since individuals benefit from individualized treatment \cite{zahabi2020adaptive}. 
Adaptive approaches generally structure this as a selection problem: choosing between pre-existing content to better match an individual's treatment needs, resulting in better health outcomes \cite{zahabi2020adaptive}. 
Current adaptive computer-assisted therapy is mainly therapist-guided or rule-based. 
The former requires therapist intervention which is time-consuming, burdensome, and requires specialized training.
The latter modifies the therapeutic content based on predefined rules, which are brittle and cannot account for all possible individuals \cite{abdessalem2017real, heloir2014design}. 
In contrast, an EDPCG framework could be used in computer-assisted therapy as an automatic content generation tool that adapts to satisfying individuals' therapeutic needs.

There are only a handful of studies that have used EDPCG for physical rehabilitation, e.g., motor rehabilitation \cite{dimovska2010towards}, amblyopia \cite{correa2014new}, and upper limb rehabilitation \cite{hocine2015adaptation}. In these studies, the Player Experience Model (PEM) in the EDPCG framework is gameplay-based (player performance). 
Gameplay-based PEM assumes that the player's internal state can be derived from the way a player plays a game. 
However, according to Yannakakis and Togelius ~\cite{yannakakis2011experience}, gameplay-based PEM is a low-resolution model of the player's experience, which is not ideal for psychological rehabilitation. Further, less attention has been paid to using EDPCG for psychological rehabilitation, e.g., mental health \cite{i2018toward}. We draw on physiological measures in our PEM, which more closely correspond to a player's internal state, especially for measuring stress. 


This paper presents an EDPCG framework for arachnophobia treatment leveraging a Experience-driven Reinforcement Learning-based PCG (EDPCGRL) content generator. The RL agent dynamically generates spiders in order to induce a desired stress level. The framework's goal is to keep players within a defined physiological state to allow for more effective exposure therapy. Exposure therapy is a therapy technique for treating anxiety disorders in which an individual is gradually exposed to the anxiety source. This paper introduces a new research area: EDPCGRL. We investigate the application of an EDPCGRL system to computer-assisted therapy. The contributions of our framework are: 1) Demonstrating the feasibility of a cognitive-based EDPCGRL approach for rehabilitation; 2) Demonstrating the first instance, to the best of our knowledge, of PCGRL outperforming search-based PCG.


\begin{table*}[tb]
\centering

\begin{tabular}{|p{2.7cm}|p{3.7cm}|p{2.7cm}|p{3cm}|p{3.5cm}|}
\hline
Study & Reward Function & Policy & Focus & Case Study \\ \hline

\cite{guzdial2018co} & Usability and helpfulness of the RL's suggestion
& Deep Q-Learning & Partner in co-creative level design &  Super Mario Bros. \\ \hline

\cite{nam2019generation} & Difficulty-based evaluation function & Deep Q-Learning, Deep Deterministic Policy Gradient & Generate diverse stages &  Implemented platforms (battle and non-battle recovery sections) \\ \hline

\cite{lopez2019reinforcement} \cite{lopez2020deep} & Efficiency and functionality of the generated layout & Proximal Policy Optimization & Placement of virtual objects & Manufacturing facility \\ \hline

\cite{cunningham2020multi} & Efficiency and functionality of the generated layout & Proximal Policy Optimization & Placement of virtual objects & Manufacturing facility, Grocery store \\ \hline

\cite{khalifa2020pcgrl} & Playable levels and satisfying some rules & Proximal Policy Optimization & 2D maps and game element placement & Binary, Zelda, Sobokon \\ \hline

\cite{gisslen2021adversarial} & Based on the solver's performance & Adversarial RL & Environment (racing tracks, platforms, paths) & Racing game \\ \hline

\cite{susanto2021applying} & Comparison between achieved and desired goal & Deep Q-Learning & Level generation & Bit-flipping, Lights out, Leaping Frog, Google chrome's Dinosaur game \\ \hline

\cite{earle2021learning} & Closeness of the generated level to target metrics & Proximal Policy Optimization & Level generation & Binary, Zelda, Sokobon, SimCity, micro-RCT \\ \hline

\end{tabular}

\caption{Summary of prior PCGRL work.}
\label{tab:table1}
\end{table*}

\section{Related Work}

\subsection{VR for Arachnophobia}
Virtual reality (VR) immerses individuals in graphical computer-generated environments. 
VR has been found to be effective in exposure therapy for specific phobias, including fear of heights \cite{freeman2018automated}, fear of spiders (Arachnophobia) \cite{cote2009cognitive} and other anxiety disorders \cite{maples2017use}. 
In exposure therapy, a subject is gradually exposed to a feared situation or object in a safe environment, leading to desensitization and a healthier response.
While we are not dismissing non-adaptive VR for arachnophobia such as \cite{shiban2015effect, miloff2019automated}, we focus on adaptive approaches, because they better fit subjects' different needs. 
For example, Kritikos et al. \cite{kritikos2021personalized} defined rules to change a spider's appearance and pattern of behaviour (e.g., size and velocity of the spider, probability of walking towards the user, etc.) to induce a desired level of anxiety in a subject. The level of anxiety is calculated based on normalized electrodermal activity (EDA~\footnote{The EDA measures the variations in the skin's electrical conductance due to increases in the activity of sweat glands}) changes. Instead of using predefined rules based on EDA changes, our framework adapts a spider using PCGRL.

\subsection{PCG}
Procedural Content Generation (PCG) automatically generates content using algorithms. Traditional PCG approaches require hand-authored knowledge, such as constructive \cite{shaker2016procedural}, search-based \cite{togelius2011search}, and constraint-based methods \cite{smith2011answer}. 
To address this limitation, researchers started applying machine learning (ML) methods to PCG \cite{summerville2018procedural}; however, because they are primarily supervised learning methods, they require a pre-existing dataset. Our framework is based on PCGRL to automatically generate new content without the need of a dataset.

\subsection{PCG for Rehabilitation}
PCG has rarely been applied to rehabilitation.
Dimovska et al. \cite{dimovska2010towards} used a constructive PCG generator in a ski-slalom game to place challenges according to a player's performance for motor rehabilitation. 
Correa et al. \cite{correa2014new} developed an adaptive first-person shooter game using constructive PCG for amblyopia treatment based on the player's performance.
Hocine et al. \cite{hocine2015adaptation} developed a game for upper limb rehabilitation through pointing tasks, i.e., reaching targets. They locate targets using Monte Carlo tree Search (MCTS) based on the user's performance, and generate the level with constructive PCG (i.e., choosing game entities).
Badia et al. \cite{i2018toward} developed a labyrinth game that adapts by estimating a user's emotions via physiological responses. The system promotes an emotional self awareness for more effective emotion regulation. A constructive PCG approach was applied to generate different graphical content. 
These works use a set of predefined content, assuming the subjects are known. Instead, our work assumes that the subjects are unknown; therefore, content needs to be generated and adapted dynamically.

\subsection{PCGRL}
PCG via Reinforcement learning (PCGRL) methods focus on applications where we do not have any pre-authored training data, but we do have an environment of possible content and a way of automatically evaluating that content. In contrast to supervised PCGML methods, no pre-existing data is required for PCGRL. Thus, it can be applied to situations like ours in which no data is available.

We summarize prior PCGRL work in Table \ref{tab:table1}. 
The purpose of these projects is mainly entertainment or education. 
The reward functions used in these works are based on different game-specific content. 
The application of a game-specific reward function, in our framework, would require assuming that the game-specific content reflects the players' cognitive state. However, the assumption is not necessarily true, and would be challenging to implement since it requires a sophisticated game design. 
Instead, based on Shaker \cite{shaker2016intrinsically}, we define our reward function via human interaction with the system, which we argue is an essential factor in effectively using PCGRL for rehabilitation purposes.

\begin{figure}[tb]
\centering
\includegraphics[width=0.9\columnwidth]{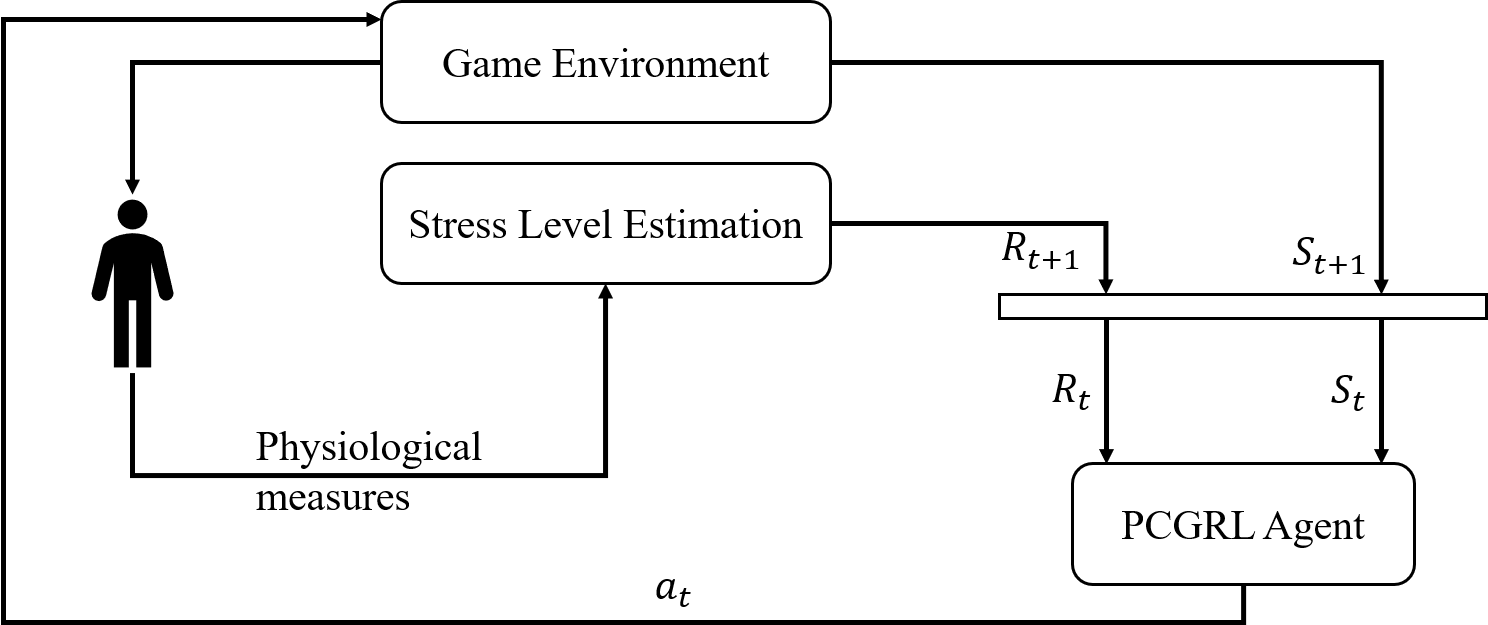}
\caption{Proposed system framework}.
\label{Fig: system}
\end{figure}

\section{Methodology}

This section overviews our framework for adapting a game environment using PCGRL for exposure therapy based on user responses. We visualize our framework in Figure~\ref{Fig: system}. 
We use arachnophobia as a case study for this framework. A subject interacts with a virtual spider, and their physiological responses are used to estimate their stress levels in real-time. A PCGRL agent modifies the virtual spider in order to reach a therapist-defined goal stress level. While we intend for the final version of this system to draw on prior work on estimating stress from physiological responses \cite{schmidt2018introducing}, we do not implement this part of the framework for this paper. In the following subsections, we describe each component of Figure~\ref{Fig: system} in detail.

\subsection{Game Environment}
For the environment, we envision a VR program with a generated 3D model of a spider. 
Our initial implementation does not require the completed VR environment and so we use a prototype version of the game environment, which is a simple representation of the spider given its specific attributes. We visualize three sets of attribute values for our prototype spider in Figure \ref{fig: Spiders}. The virtual spider is initially generated with random attributes, which are later adjusted by our EDPCG approach. These attributes include movement-related and appearance characteristics of spiders based on Lindner et al.'s work \cite{lindner2019so}. In this study, spider-fearful individuals ($n = 194$) were asked to rate the characteristics of spiders based on their fear response. We overview this study in more detail in the next subsection. 

\begin{table}[t]
\resizebox{\columnwidth}{!}{\begin{tabular}{|l|l|l|l|}
\cline{2-4}
\multicolumn{1}{l|}{} & \multicolumn{1}{l|}{Attribute} & \multicolumn{1}{l|}{Impact Factor} & \multicolumn{1}{l|}{Possible Values} \\ \hline
\multicolumn{1}{|l|}{\multirow{3}{*}{\rotatebox[origin=c]{90}{Movement Factors}}} & \multicolumn{1}{l|}{(1) Locomotion} & \multicolumn{1}{l|}{\begin{tabular}[c]{@{}l@{}} $\mu: 0.9$ \\ $\delta: 0.15$ \end{tabular}} & \multicolumn{1}{l|}{\begin{tabular}[c]{@{}l@{}}0: Standing\\ 1: Human-like locomotion\\ 2: Spider-like locomotion\end{tabular}} \\ \cline{2-4} 
\multicolumn{1}{|l|}{} & \multicolumn{1}{l|}{\begin{tabular}[c]{@{}l@{}} (2) Amount \\ of movement\end{tabular}} & \multicolumn{1}{l|}{\begin{tabular}[c]{@{}l@{}} $\mu: 0.9$ \\ $\delta: 0.15$ \end{tabular}} & \multicolumn{1}{l|}{\begin{tabular}[c]{@{}l@{}}0: Slightly   \\ 1: Moderate  \\ 2: Too much\end{tabular}} \\ \cline{2-4} 
\multicolumn{1}{|l|}{} & \multicolumn{1}{l|}{(3) Closeness} & \multicolumn{1}{l|}{\begin{tabular}[c]{@{}l@{}} $\mu: 0.4$ \\ $\delta: 0.17$ \end{tabular}} & \multicolumn{1}{l|}{\begin{tabular}[c]{@{}l@{}}0: Far away\\ 1: In the middle\\ 2: Very close\end{tabular}} \\ \hline
\multicolumn{1}{|l|}{\multirow{3}{*}{\rotatebox[origin=c]{90}{Appearance Factors}}} & \multicolumn{1}{l|}{(4) Largeness} & \multicolumn{1}{l|}{\begin{tabular}[c]{@{}l@{}} $\mu: 0.7$ \\ $\delta: 0.16$ \end{tabular}} & \multicolumn{1}{l|}{\begin{tabular}[c]{@{}l@{}}0: Small\\  1: Medium\\  2: Large\end{tabular}} \\ \cline{2-4} 
\multicolumn{1}{|l|}{} & \multicolumn{1}{l|}{(5) Hairiness} & \multicolumn{1}{l|}{\begin{tabular}[c]{@{}l@{}} $\mu: 0.6$ \\ $\delta: 0.21$ \end{tabular}} & \multicolumn{1}{l|}{\begin{tabular}[c]{@{}l@{}}0: Without hair\\ 1: With hair\end{tabular}} \\ \cline{2-4} 
\multicolumn{1}{|l|}{} & \multicolumn{1}{l|}{(6) Color} & \multicolumn{1}{l|}{\begin{tabular}[c]{@{}l@{}} $\mu: 0.5$ \\ $\delta: 0.20$ \end{tabular}} & \multicolumn{1}{l|}{\begin{tabular}[c]{@{}l@{}}0: Gray\\ 1: Brown\\ 2: Black\end{tabular}} \\ \hline
\end{tabular}}
\caption{List of the Attributes and their Impact factors on fear (mean and standard deviation) based on ~\cite{lindner2019so}. We define ordinal possible values for each attribute.}
\label{Table: Attributes}
\end{table}

\subsection{Stress Level Estimation}
Theoretically, players' physiological responses should allow for an estimation of their stress level~\cite{schmidt2018introducing}.
The estimated stress level is compared with a target stress level (explicitly identified for exposure therapy purposes by a therapist) to generate reward.
Given the ongoing global health crisis, we calculate the stress levels of virtual subjects as functions of the spider's attributes in order to develop a simplified version of our framework as a proof of concept. 
Consequently, we generate various virtual subjects, i.e., each has its own stress level function corresponding to spider attributes, based on the Lindner et al. study~\cite{lindner2019so}. 
The paper introduced seven spider attributes and measured their impact on their participants' overall amount of fear by asking the participants to quantitatively rate each spider attribute. We choose six attributes for spider generation as shown in Table~\ref{Table: Attributes}. We drop the realness attribute since it indicates whether or not to use real spiders, and we focus on virtual spiders. For each attribute, we define 2-3 ordinal values. 

The movement attributes: \emph{locomotion}, \emph{amount of movement} and \emph{closeness}, denote how the spider moves (specifically the movement of the legs), how much it moves, and how close it gets to the subject respectively. The appearance attributes: \emph{largeness}, \emph{hairiness}, and \emph{color}, denote the size of the spider,  whether the spider is hairy or not, and the color of the spider, respectively.

We generate 100 virtual subjects as stress estimation functions, each with different responses to spider attributes.  In~\cite{lindner2019so}, an impact factor is associated with each spider attribute, denoting the effect of the attribute on the subjects' amount of fear.
Based on the impact factors reported in the study, we model a normal distribution for each of the attributes as
$
N\sim (\mu_{a_i} , \delta_{a_i}) , i \in \left \{ 1,...,6 \right \}
$;
where $a_i$ denotes the $i$-th attribute in Table~\ref{Table: Attributes}, and $\mu_{a_i}$ and $\delta_{a_i}$ are derived from the $i$-th attribute's mean impact factor and subjects' fear variance of the attribute, respectively. 
We draw 100 random samples from each distribution to generate 100 virtual humans as our subjects. 
An example of stress for one virtual subject is: $
1.37 \times ( 0.97 \times a_1 + 0.87 \times a_2 +
0.07 \times a_3 + 0.63 \times a_4 +
0.67 \times a_5 + 0.77 \times a_6 )
$; Where $1.37$ is the coefficient to scale the stress level to the range 0-10 for this virtual subject. 
This approach led to 100 unique virtual stress levels, which are still based on psychological study of real, spider-fearing humans. Thus, we expect there to be at least some psychological grounding to our results.


\subsection{Markov Decision Problem}
\subsubsection{State}
Each state represents a combination of the spider's attributes. We define for each attribute a range of possible values, which are listed in Table~\ref{Table: Attributes}. The representation of state in time $t$ can be represented as (Eq.~\ref{Eq: state}):
\begin{equation}
\label{Eq: state}
{S_{t}} = \left \{ a_{i,t} \right \} , i \in \left \{ 1,...,6 \right \}
\end{equation}
\noindent Where $a_{i,t}$ is the value of the $i$-th attribute in time $t$.

\subsubsection{Action}
The action is defined as increasing or decreasing one attribute at a time (Eq.~\ref{Eq: action}). 
\begin{equation}
\label{Eq: action}
A_{t} = \left \{ a_{policy, t} \pm 1 \right \}
\end{equation}
\noindent Where $a_{policy,t}$ is the attribute chosen by the policy of EDPCGRL for action in time $t$. The intuition behind this decision is that therapists gradually increase/decrease these attributes in exposure therapy to find reasonable values to produce their intended stress level. 

\subsubsection{Reward}
The reward function is calculated based on a normal distribution $N(\mu, \delta)$, where $\mu$ is the target stress level, and $\delta = (MaxStress - MinStress) / 2$, where $MaxStress$ and $MinStress$ are maximum and minimum values that the stress level can reach respectively. 
We scale the distribution to the range of $(-1, 1)$ such that the target stress level achieves a reward of 1, and the reward decreases as the stress level gets further from the target stress. 
The resulting reward function is shown in Eq.~\ref{Eq: reward}.

\begin{equation}
\label{Eq: reward}
\begin{split}
Reward = \frac
{2 e^{-0.5 (\frac{x-\mu }{\delta})^{2}} - e^{-0.5 (\frac{\alpha -\mu }{\delta})^{2}} - 1}
{1 - e^{-0.5 (\frac{\alpha -\mu }{\delta})^{2}}}   \\
\alpha =\begin{cases}
    MaxStress, & \text{if $\mu < \delta$}.\\
    MinStress, & \text{if $\mu \geq \delta$}.
  \end{cases}
\end{split}
\end{equation}

\noindent where $x$ is the current stress level.
\\

\section{Evaluation}

\begin{figure*}[t]
  \centering
  \begin{subfigure}[b]{0.266\textwidth}
    \includegraphics[width=\textwidth]{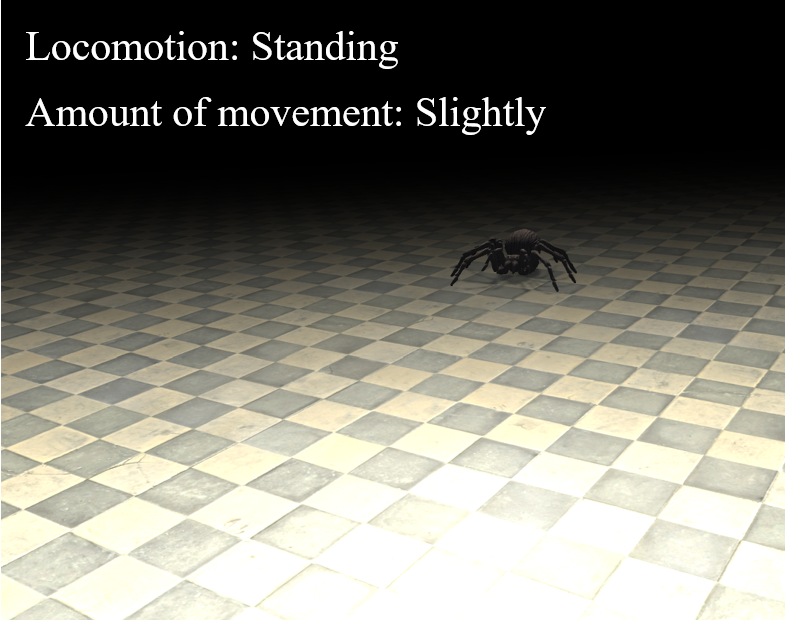}
     \caption{Minimum}
  \end{subfigure}
  \begin{subfigure}[b]{0.266\textwidth}
    \includegraphics[width=\textwidth]{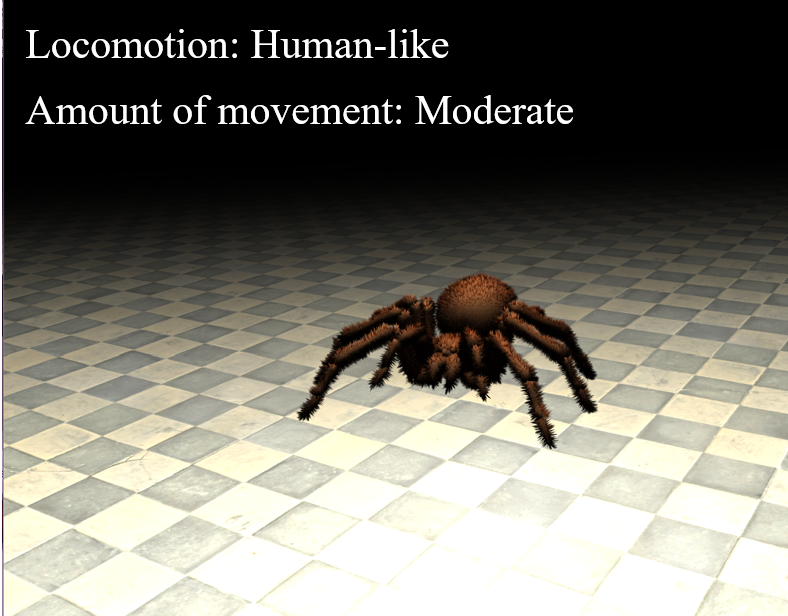}
    \caption{Average}
  \end{subfigure}
  \begin{subfigure}[b]{0.266\textwidth}
    \includegraphics[width=\textwidth]{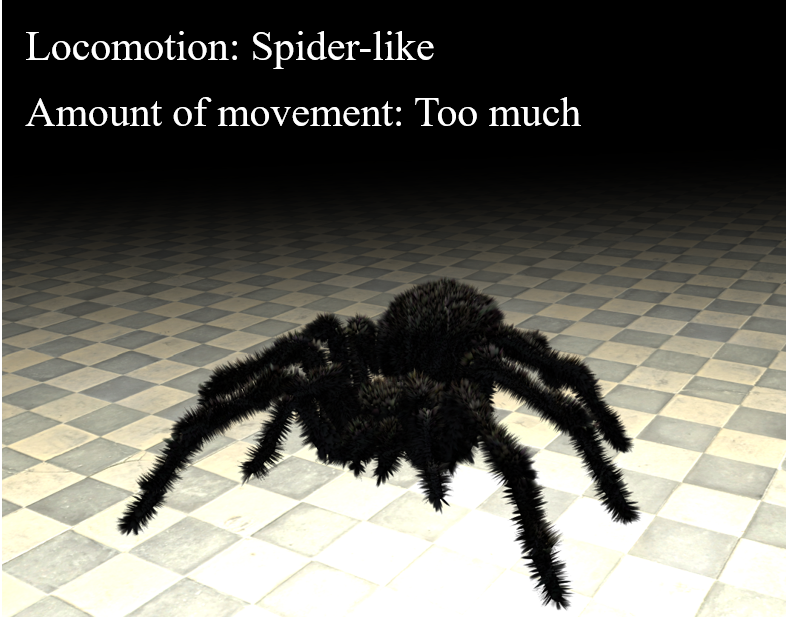}
    \caption{Maximum}
  \end{subfigure}
  \caption{Generated spiders with three initial attributes from our initial prototype.}
  \label{fig: Spiders}
\end{figure*}

We propose a framework for arachnophobia exposure therapy that can automatically adapt to subjects. For each subject, it generates spiders until it finds a spider (with specific features) that induces a subject's target stress level. 
This is a difficult task since it is suggested in~\cite{vetter2013arachnophobic} that people with arachnophobia respond to different aspects of spiders differently.
We evaluate our proposed method in terms of how many times we present new spiders to a subject, denoted as \emph{Spiders Presented}. 
This number should be as low as possible, because it may adversely affect the effectiveness of the treatment. This metric is equivalent to the number of times that an algorithm outputs a new spider.

At this stage, we evaluate our framework using virtual subjects (discussed in the methodology section) as a proof of concept. This is because of the ongoing COVID-19 pandemic, which stopped us from conducting a human subject study. We generate 100 virtual subjects with a theoretical basis in real spider-fearing individuals. If our method outperforms our baseline methods using these virtual subjects, it will indicate that we can likely expect similar performance with real subjects. 

Since there is no existing EDPCGRL framework to compare against, we compare our proposed method to EDPCG via search-based procedural content generation (SBPCG) methods. SBPCG is a traditional way to adapt content based on the assessment of users \cite{risi2015petalz, liapis2013sentient}. We briefly describe the baseline methods used in our evaluation and the implementation of our proposed method.

\subsubsection{Genetic Algorithm:}
A Genetic Algorithm (GA) is our first baseline because it is a popular experience-driven SBPCG approach~\cite{togelius2011search}. In our problem, the GA's chromosome is equal to the state defined in the previous section. The GA starts with a population that consists of a given initial state and its nine neighbour states (the population size is 10).
We denote two states as neighbours if and only if they differ by only 1 value for a single attribute. 
Our approach selects the ten best chromosomes in terms of their fitness as its initial population.
Based on Table~\ref{Table: Attributes}, there are two chromosomes that could have eleven neighbours (locomotion=1, amount of movement=1, closeness=1, largeness=1, hairiness=0 or 1, color=1). 
The "average" initial state is one of these chromosomes. Therefore, we do not anticipate different results if we took all available neighbours. 
The GA then performs crossover, mutation, and selection until it reaches the termination condition. The crossover process uses a weighted sampling for picking two pairs of chromosomes based on fitness score. The pairs are swapped from the middle point (one child takes the first half of the attributes from one parent and the second half from the other, and the other child vice versa) to generate two new offspring. For mutation, a random attribute in the new offspring is changed with $0.1$ probability to a random, valid. These operations do not hold the constraint of increasing/decreasing one attribute at a time, allowing for much larger steps through the search space. Finally, the selection process chooses ten chromosomes with the highest fitness values as the next population. 

\subsubsection{Greedy search:} 
Our second baseline is a simple greedy search. 
This approach works best if the fitness function does not have local optima since it always exploits the best neighbour. It starts from the given initial state, searches its neighbours, and chooses the neighbour with the maximum fitness. Therefore, the fitness gradually increases (or stays steady) until the termination condition. This approach should perform best in terms of the minimum number of \emph{Spiders Presented} if our problem is simple enough. 

\subsubsection{Random Search:} 
Our final baseline is a random search. It also starts from the same initial state as previous baselines. In each step, it randomly selects an action (defined in the previous section) and goes to a new state. It keeps searching until it reaches the termination condition. We include random search to investigate the importance of exploration in this domain.

\subsubsection{PCGRL (Our method):}
This is our method, which optimizes the spiders' attributes using RL.
We employ the tabular Q-learning algorithm and epsilon-greedy ($\epsilon = 0.05$) for our action selection policy~\cite{sutton2018reinforcement}.
We initialize a Q-table with either random or zero values that store the values of state-action pairs, and update the values in each iteration.
\\
\\
These methods all use the same fitness function: the reward function for the PCGRL agent. We use the same initial state and termination criteria for all the methods for a fair evaluation. We define these as:
\begin{itemize}
    \item Initial State: We use three initial states for our evaluation: in each state, we set the spider's attributes to either the maximum, minimum, or the average values within each attribute's range (visualized in Figure \ref{fig: Spiders}).
    \item Termination: These algorithms terminate if they reach one of these criteria: achieve a state with maximum fitness or run for 100 iterations. 
\end{itemize}

We run each algorithm 10 times for each virtual subject and for each of the three initial states. We apply these approaches for every goal stress level from 1 to 9. In total, each algorithm is executed 27000 times.

\section{Result}

\begin{table*}[t]
\begin{tabular}{|p{0.7cm}|p{1.2cm}|l|l|l|l|l|l|}
\Xhline{3\arrayrulewidth}
\begin{tabular}[c]{@{}l@{}}Initial \\ State\end{tabular} & Stress & Metric & Random & Greedy & GA & RL\_Random & RL\_Zero \\ \Xhline{3\arrayrulewidth}
\multirow{6}{*}{Min} & \multirow{2}{*}{Low} & Spiders Presented & 5.32$\pm$7.75\oneS & 11.24$\pm$1.58 & 9.94$\pm$1.76 & 6.17$\pm$8.69 & \textbf{4.04$\pm$2.08}\oneS \\ \cline{3-8} 
 &  & Accuracy & 99.26 & 100 & 100 & 99.33 & 99.90 \\ \cline{2-8} 
 & \multirow{2}{*}{Moderate} & Spiders Presented & 11.15$\pm$5.24 & 21.95$\pm$2.48 & 25.27$\pm$5.66 & \textbf{8.43$\pm$3.70}\twoS & 11.37$\pm$4.84 \\ \cline{3-8} 
 &  & Accuracy & 100 & 99 & 100 & 99.90 & 97.83 \\ \cline{2-8} 
 & \multirow{2}{*}{High} & Spiders Presented & 39.86$\pm$18.81 & 37.38$\pm$4.14 & 49.09$\pm$10.31 & \textbf{33.10$\pm$17.72}\twoS & 29.04$\pm$9.83 \\ \cline{3-8} 
 &  & Accuracy & 81.90 & 63.33 & 100 & 89.76 & 53 \\ \Xhline{3\arrayrulewidth}
\multirow{6}{*}{Avg} & \multirow{2}{*}{Low} & Spiders Presented & 32.06$\pm$19.66 & 27.38$\pm$1.49 & 29.82$\pm$5.97 & \textbf{19.17$\pm$13.52}\twoS & 16.83$\pm$9.17 \\ \cline{3-8}
 &  & Accuracy & 83.6 & 96 & 100 & 93.06 & 72.63 \\ \cline{2-8} 
 & \multirow{2}{*}{Moderate} & Spiders Presented & 5.11$\pm$6.63\oneS & 8.33$\pm$3.23 & 10$\pm$0 & 5.62$\pm$5.39 & \textbf{3.95$\pm$3.66}\oneS \\ \cline{3-8} 
 &  & Accuracy & 100 & 100 & 100 & 100 & 99.06 \\ \cline{2-8} 
 & \multirow{2}{*}{High} & Spiders Presented & 30.42$\pm$20.17 & \textbf{20.57$\pm$1.71}\twoS & 23.34$\pm$5.78 & 23.50$\pm$13.15 & 22.50$\pm$11.00 \\ \cline{3-8} 
 &  & Accuracy & 85.5 & 97.66 & 100 & 90.93 & 52.93 \\ \Xhline{3\arrayrulewidth}
\multirow{6}{*}{Max} & \multirow{2}{*}{Low} & Spiders Presented & 38.01$\pm$18.67 & 37.40$\pm$4.14 & 50.05$\pm$10.31 & \textbf{24.77$\pm$15.16}\twoS & 22.51 (8.49) \\ \cline{3-8} 
 &  & Accuracy & 81.30 & 63.66 & 100 & 93.06 & 70.93 \\ \cline{2-8} 
 & \multirow{2}{*}{Moderate} & Spiders Presented & 11.45$\pm$5.27 & 23.95$\pm$2.48 & 25.24$\pm$6.38 & 9.29$\pm$4.49\oneS & \textbf{8.56$\pm$3.32}\oneS \\ \cline{3-8} 
 &  & Accuracy & 100 & 99 & 100 & 99.93 & 99.40 \\ \cline{2-8} 
 & \multirow{2}{*}{High} & Spiders Presented & 5.39$\pm$8.52\oneS & 11.24$\pm$1.58 & 9.88$\pm$1.72 & 4.83$\pm$5.30\oneS & \textbf{4.20$\pm$5.32}\oneS \\ \cline{3-8} 
 &  & Accuracy & 99.43 & 100 & 100 & 99.20 & 95.90 \\ \Xhline{3\arrayrulewidth}
\end{tabular}
\caption{Mean and standard deviation for our Spiders Presented metric and the Accuracy of each approach across initial states and goal stress values.}
\label{Table:result_category}
\end{table*}

We define three stress categories: low ($[1, 3]$), moderate ($[4, 6]$), and high ($[7, 9]$). We calculate the average value of \emph{Spiders Presented} for our 100 virtual subjects for each initial state and each stress category. We repeat this process 10 times and report the averages in Table~\ref{Table:result_category}.  

\emph{Spiders Presented} in the table denotes the number of new spiders showed to a virtual subject on average when the approach is successful. The \emph{Accuracy} shows the percentage of times the method could find a spider that provokes the target stress level out of the ten attempts on average. The \emph{Spiders Presented} result is not taken into account if a run had \emph{Accuracy} less than 75\%.
For example, for the \emph{Max} initial state and \emph{Low} stress goal, the \emph{RL\_Zero} has the least number of Spiders Presented, but since the Accuracy is 70.93\% (less than 75\%), we did not consider it to be the best result.
In each row, the method that outperforms other methods according to the \emph{Spiders Presented} metric is shown with bold text. The table show that our proposed method outperformed the baseline methods for almost all of the target stress levels with various initial states.
We note that we do not expect these results to necessarily generalize to all hyperparameter values.
A hyperparameter sweep of the approaches is not in the scope and within the page limitation of this paper.

Two-tailed, paired-samples t-tests ($p<0.05$) were performed to compare the mean \emph{Spiders Presented} metric across different methods. The results are shown in the table with \emph{\twoS} if the best algorithm is significantly better than the other methods. There are cases that the best method significantly differs from the others except for one/two method(s). In these cases, we indicate the method with \emph{\oneS}.

\section{Discussion}

\begin{figure*}[t]
\centering
\includegraphics[width=0.8\textwidth]{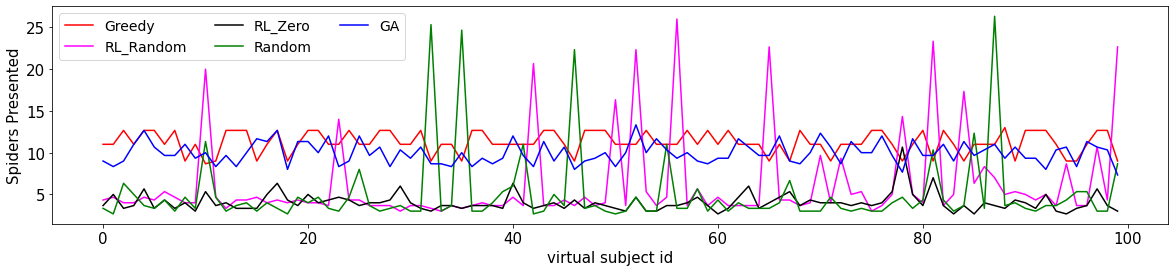}
\caption{Spiders Presented for each virtual subject for the minimum initial state with target stress level set to low.}
\label{Fig: initialMin_StressMin}
\end{figure*}

We found that our proposed method, EDPCGRL, outperforms the baseline methods in the \emph{Spiders Presented} metric, i.e., PCGRL showed fewer spiders to our virtual subjects before finding one that induces the specific subject's stress level on average. This might be because it combines exploitation and exploration. Figure~\ref{Fig: initialMin_StressMin} shows the \emph{Spiders Presented} for each virtual subject in the sequence they were presented to all approaches. 
It reveals that the RL agent with the q-table initialized to 0 consistently outperforms other methods, which shows the method did not just learn our subjects' behaviour over time.

Our results reveal that for each stress level category: low, moderate, and high, it might be ideal to set the initial state attributes to the minimum, average, and maximum, respectively.
However, we emphasize that increasing/decreasing all the spider's attributes does not always produce spiders closer to our desired one due to the variance in the virtual subject fitness functions. Nevertheless, in these situations, the Random baseline performs near optimally, because it is already closer to the desired spider in most cases. 

We evaluate these methods on the virtual subjects, which simplifies the problem since the subjects are deterministic and do not change over time. However, if we imagine our virtual subjects not as distinct individuals, but as the same subject over time (e.g. fig. \ref{Fig: initialMin_StressMin}), we can observe that RL would also likely do better with dynamic, real-world individuals. Therefore, our results show that our method has potential for real subjects. 

We intend to evaluate the proposed method on real human subjects who are complicated and may respond differently over time. Therefore, it is challenging to obtain a model that maps physiological measures to stress levels. In fact, the same individuals may show different patterns in their physiological measures. Environmental factors may affect individuals' physiological response, e.g., drinking coffee, sleeping less/more. We require a robust and reliable model in our future work that accurately estimates stress levels from physiological measures. We also plan to improve the fidelity of the virtual spiders by using more sophisticated methods, such as more attributes and/or more possible ordinal values. 

Currently, our approach uses a simple yet effective RL method. However, we are interested in investigating more sophisticated RL methods to determine if their performance differs significantly.
There are many other algorithms available that utilize the combination of exploitation and exploration, e.g.,  Monte Carlo Tree Search (MCTS).  However, they would require hyperparameter finetuning to work properly in our domain. 
We are also interested in applying transfer learning approaches to adapt the knowledge learned from one subject to a new subject.

Another limitation of our work is related to the ethical aspects. Our framework might cause excessive stress in the subjects. Therefore, the action selection policy in our PCGRL should be carefully adjusted in a way that the RL agent considers gradually increasing the stress level. Nevertheless, the framework might also re-traumatize the subjects.


\section{Conclusion}
This paper introduces a new research area, i.e., EDPCGRL which was unidentified in prior work. We defined a proof of concept of our EDPCGRL framework for virtual reality exposure therapy, where PCGRL adjusts game content according to a subjects' physiological measures. We hypothesized that our EDPCG framework could be applied in computer-assisted therapy rather than reliant on pre-authored methods. We support this hypothesis by evaluating our proposed framework for arachnophobia in a case study. Different spiders with six different attributes are generated based on a subject's stress level to find one that induces a target stress level. Our goal was to design a method that finds a desired spider with fewer spiders presented to a subject. We found that EDPCGRL outperformed existing experience-driven SBPCG methods for this task.
\section{Acknowledgements}

We acknowledge the support of the Natural Sciences and Engineering Research Council of Canada (NSERC) and CISCO systems.

\begin{small}
\bibliography{Main}
\end{small}

\end{document}